\documentclass[letterpaper]{article} % DO NOT CHANGE THIS
\usepackage{aaai25}  % DO NOT CHANGE THIS
\usepackage{times}  % DO NOT CHANGE THIS
\usepackage{helvet}  % DO NOT CHANGE THIS
\usepackage{courier}  % DO NOT CHANGE THIS
\usepackage[hyphens]{url}  % DO NOT CHANGE THIS
\usepackage{graphicx} % DO NOT CHANGE THIS
\usepackage{natbib}  % DO NOT CHANGE THIS AND DO NOT ADD ANY OPTIONS TO IT
\usepackage{caption} % DO NOT CHANGE THIS AND DO NOT ADD ANY OPTIONS TO IT
\usepackage{algorithm}
\usepackage{algorithmic}
\usepackage{booktabs}
\usepackage{amsmath}
\usepackage{amssymb}
\usepackage{multirow}
\usepackage [switch]{lineno}
\usepackage{amsfonts}
\usepackage{newfloat}
\usepackage{listings}
\DeclareCaptionStyle{ruled}{labelfont=normalfont,labelsep=colon,strut=off} % DO NOT CHANGE THIS
\floatstyle{ruled}
\newfloat{listing}{tb}{lst}{}
\floatname{listing}{Listing}
\title{DAPoinTr: Domain Adaptive Point Transformer for Point Cloud Completion}
\author{
    Yinghui Li \textsuperscript{\rm 1}\equalcontrib,
    Qianyu Zhou \textsuperscript{\rm 2}\equalcontrib,
    Jingyu Gong \textsuperscript{\rm 3},
    Ye Zhu\textsuperscript{\rm 1},
    Richard Dazeley\textsuperscript{\rm 1},\\
    Xinkui Zhao\textsuperscript{\rm 4}\thanks{Corresponding Authors.},
    Xuequan Lu\textsuperscript{\rm 5}\footnotemark[2]
}
\affiliations{
    \textsuperscript{\rm 1}School of Information Technology, Deakin University, Australia;\\
    \textsuperscript{\rm 2}Department of Computer Science and Engineering, Shanghai Jiao Tong University, Shanghai, China;\\
    \textsuperscript{\rm 3}School of Computer Science and Technology, East China Normal University, Shanghai, China;\\
    \textsuperscript{\rm 4}School of Software Technology, Zhejiang University, Ningbo, China;\\
    \textsuperscript{\rm 5}Department of Computer Science and Software Engineering, The University of Western Australia, Australia.\\
    s218161821@deakin.edu.au, zhouqianyu@sjtu.edu.cn, jygong@cs.ecnu.edu.cn, ye.zhu@deakin.edu.au, richard.dazeley@deakin.edu.au, zhaoxinkui@zju.edu.cn, bruce.lu@uwa.edu.au

}

\usepackage{bibentry}
\begin{document}
\maketitle

\begin{abstract}
Point Transformers (PoinTr) have shown great potential in point cloud completion recently. Nevertheless, effective domain adaptation that improves transferability toward target domains remains unexplored. In this paper, we delve into this topic and empirically discover that direct feature alignment on point Transformer's CNN backbone only brings limited improvements since it cannot guarantee sequence-wise domain-invariant features in the Transformer. To this end, we propose a pioneering Domain Adaptive Point Transformer (DAPoinTr) framework for point cloud completion. DAPoinTr consists of three key components: Domain Query-based Feature Alignment (DQFA), Point Token-wise Feature alignment (PTFA), and Voted Prediction Consistency (VPC). In particular, DQFA is presented to narrow the global domain gaps from the sequence via the presented domain proxy and domain query at the Transformer encoder and decoder, respectively. PTFA is proposed to close the local domain shifts by aligning the tokens, \emph{i.e.,} point proxy and dynamic query, at the Transformer encoder and decoder, respectively. VPC is designed to consider different Transformer decoders as multiple of experts (MoE) for ensembled prediction voting and pseudo-label generation. Extensive experiments with visualization on several domain adaptation benchmarks demonstrate the effectiveness and superiority of our DAPoinTr compared with state-of-the-art methods. Code will be publicly available at: https://github.com/Yinghui-Li-New/DAPoinTr
\end{abstract}


\section{Introduction}
3D point cloud data plays a pivotal role across diverse domains, including autonomous vehicles, robotics, augmented and virtual reality, \emph{etc}. Point cloud completion (PCC) aims to predict the complete geometric shape given a partial input. Previous works~\cite{yang2018foldingnet,tchapmi2019topnet,Huang_2020_CVPR,xie2020grnet,yu2021pointr,yu2023adapointr} have significantly enhanced the performance of fully-supervised PCC models, using training datasets composed of pairs of partial and complete point clouds. 
Recently, PoinTr \cite{yu2021pointr} adopted a Transformer encoder-decoder architecture and has reformulated PCC as a set-to-set translation problem, showing excellent performance within the same domains. However, %a significant challenge arises 
when transferring these models from source domains to target domains,  they would suffer from substantially degraded performance on target datasets. This is mainly due to the existing domain gaps, \emph{e.g.,} different occlusions, viewpoints, sensor resolutions, and light reflection, across different domains.

To address this issue, unsupervised domain adaptation (UDA) techniques are introduced into point cloud completion to enhance the transferability toward real domains. This involves several categories: adversarial learning~\cite{chen2019unpaired,zhang2021unsupervised}, data reconstruction~\cite{wen2021cycle4completion}, disentangled learning~\cite{gong2022optimization}, self-supervised learning~\cite{hong2023acl}. Nevertheless, a common limitation of these UDA PCC methods is their reliance on Convolution Neural Networks (CNNs) and suffer from limited receptive fields as layers are stacked, which can be problematic for capturing the broader context required for learning domain-agnostic features.  With the recent surge of point Transformers, is it possible to empower them with such a capability to perform accurate completion in cross-domain scenarios?

\begin{figure*}
    \centering
    %\vspace{-12mm}
    \includegraphics[width=1.0\textwidth]{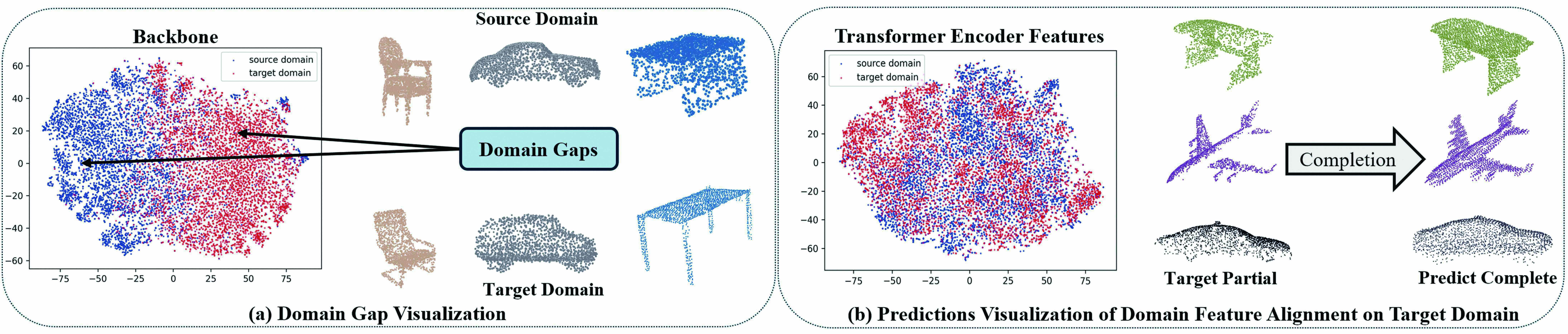}
    %\vspace{-6mm}
    \captionof{figure}{(a) Visualization of the domain discrepancy among objects of the same category from different domains where there are significant variations in topology, geometric patterns, and feature distribution.
    We observe that directly applying adversarial alignment to CNN backbone features of PoinTr ~\cite{yu2021pointr} does not well align the source domains and target domains since it does not ensure learning domain-invariant sequence features of the point cloud.
    (b) In contrast, we propose a Domain Adaptive Point Transformer (DAPoinTr) framework for point cloud completion that well aligns the cross-domain distributions (visualization of Transformer encoder features) and manages to generate complete shapes of partial point cloud input. 
    % Presents visualization of aligned feature distribution and objects visualization based on our proposed feature alignment strategies.
    }
    \label{fig1}
%\vspace{2mm}
\end{figure*}

A straightforward idea to address this issue is to directly apply adversarial alignment to features extracted by the DGCNN~\cite{phan2018dgcnn} backbone (referred to as DA-DGCNN) of PoinTr~\cite{yu2021pointr}. As illustrated in Figure~\ref{fig1}(a), such a manner does enhance the point Transformer's cross-domain performance, but the improvement is very limited when transferring from CRN~\cite{wang2020cascaded} to ModelNet~\cite{wu20153d}. 
We deduce that this limitation arises because aligning feature distributions on the DGCNN backbone does not ensure sequence-wise domain-invariant features in the subsequent Transformer, which are essential for accurate completion. Thus, the sequence features from the Transformer encoder remain domain-separated and not well-aligned, leading to less decent performance of the point Transformers due to these shifted sequence features. In Figure~\ref{fig1}(a), we also provide the visualization of the domain discrepancy among objects of the same category from different domains where there are significant variations in topology and geometric patterns.

Motivated by the aforementioned analysis, we introduce a pioneering framework, namely Domain Adaptive Point Transformer (DAPoinTr), for the point cloud completion task. DAPoinTr aims to align the sequence-wise point features with Transformer encoder-decoder architecture to enhance the adaptability toward the real target domains. It consists of three key components: Domain Query-based Feature Alignment (DQFA), Point Token-wise Feature alignment (PTFA), and Voted Prediction Consistency (VPC). Specifically, DQFA is presented to narrow the global domain gaps in layout and inter-sketch relationships from the sequence via the presented domain proxy and domain query at the Transformer encoder and decoder, respectively. A domain proxy is designed to encode global features that reflect object layout and aggregate domain-specific features from the whole sequence. Besides, the domain query encodes some sketch relationships for adaptation and fuses context features from each dynamic query in the sequence. Secondly, PTFA is proposed to close the local domain shifts by aligning the token at the Transformer encoder and decoder, respectively.
Concretely, Point Proxy Feature Alignment is presented to align each token in the encoder sequence, \emph{i.e.,} point proxy, to alleviate domain gaps caused by local shape, appearance, \emph{etc}. Besides, Dynamic Query Feature Alignment is proposed to align 
each token in the decoder sequence \emph{i.e.,} dynamic query, and to close domain gaps at the object level. Finally, VPC is designed to consider different Transformer decoders as multiple of experts (MoE) to vote for the ensembled prediction, alleviating the domain shifts between the source and target domain. Since predictions from different experts performed differently, we use the voted prediction consistency result as the threshold value for selecting qualified predictions to generate pseudo-labels to further improve the robustness and transferability.

In summary, our contributions are three-fold:
\begin{itemize}
\item We present a pioneering framework, Domain Adaptive Point Transformer (DAPoinTr) for the point cloud completion task. To the best of our knowledge, this is the first work that studies the transferability of the point Transformer
(PoinTr) in DA point cloud completion.

\item We present Domain Query-based Feature Alignment and Point Token-wise Feature Alignment for aligning sequence features on a global and local level, respectively.
Besides, we design a Voted Prediction Consistency to further regularize the sequence features and improve the discriminability of the point transformers.

\item Extensive experiments with visualization on several challenging DA benchmarks, \emph{e.g.,} KITTI, ScanNet, and MatterPort for real-world scans, and 3D-FUTURE and ModelNet for synthetic data, demonstrate our method's superiority compared with other state-of-the-art methods. 
\end{itemize}

\begin{figure*}[t]
\centering
\includegraphics[width=1.0\textwidth]{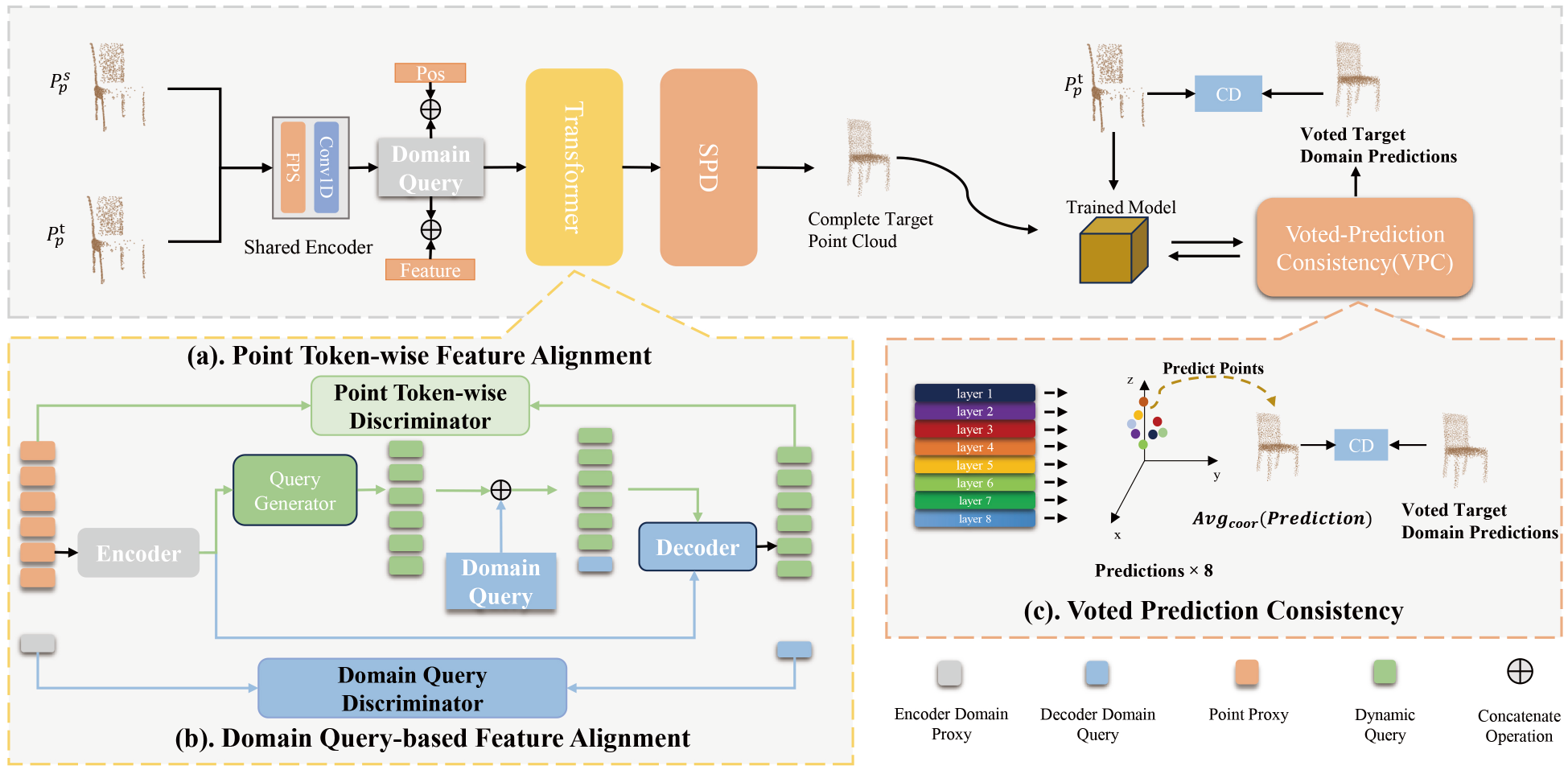} % Reduce the figure size so that it is slightly narrower than the column.
\caption{The framework of DAPoinTr for domain adaptive point cloud completion, including three key components: (a) Point Token-wise Feature Alignment (PTFA) is proposed to close the local domain shifts by aligning the tokens, \emph{i.e.,} point proxy and dynamic query, at the Transformer encoder and decoder, respectively. (b) Domain Query-based Feature Alignment (DQFA) is presented to narrow the global domain gaps from the sequence via the presented domain proxy and domain query at the Transformer encoder and decoder, respectively.  Finally, Voted Prediction Consistency (VPC) is designed to consider different Transformer decoders as multiple of experts (MoE) to vote for the ensembled prediction and pseudo-label generation.
}
\label{fig2}
\end{figure*}

\section{Related Work}
\noindent \textbf{Point Cloud Completion (PCC).} 
PCC aims to generate complete shapes from partial point clouds. 
PCN \cite{yuan2018pcn} was proposed to use the FoldingNet~\cite{yang2018foldingnet} to transform 2D points into 3D surfaces by emulating the deformation process of a 2D plane. Based on this, many techniques \cite{tchapmi2019topnet, xie2020grnet,xiang2021snowflakenet,wang2021voxel} have been developed, focusing on achieving higher resolution and enhanced robustness in PCC. For instance, AtlasNet~\cite{groueix2018papier} and 3D-EPN~\cite{Dai_2017_CVPR} incorporated convolutional neural networks (CNNs) to reconstruct complete shapes. However, they suffer from limited receptive fields as the layers are deeply stacked. To tackle this, SnowFlakeNet~\cite{xiang2021snowflakenet} was presented by conceptualizing PCC as a progressive growth from specific parent points in a snowflake-like manner, allowing for the more organic development of global structures.  Despite their gratifying progress, most of them struggle to alleviate domain gaps when applying them to real-world target domains.

\noindent \textbf{Domain Adaptation for Point Cloud Completion.}
% Rewrite
Unsupervised domain adaptation (UDA)~\cite{chen2019unpaired,jiang2024pcotta} has recently been introduced into PCC to narrow domain shifts and improve the adaptability toward target domains. The mainstream UDA PCC methods aim to learn domain-invariant features and are mainly categorized into adversarial learning~\cite{chen2019unpaired,zhang2021unsupervised}, data reconstruction~\cite{wen2021cycle4completion}, disentangled learning~\cite{gong2022optimization}, self-supervised learning~\cite{hong2023acl}, and dual mixup consistency regularized learning~\cite{liu2024cloud}. Nonetheless, these approaches heavily rely on Convolution Neural Networks (CNNs) and tend to learn local features, which can be problematic for capturing the broader global context required for learning domain-agnostic features. Thus, it is in urgent need to investigate UDA PCC with global modeling and high transferability based on vision Transformers.

\noindent \textbf{Vision Transformers.} Transformers~\cite{vaswani2017attention} have proven to be very effective in modeling long-range dependencies and have been widely explored in various vision tasks~\cite{dosovitskiy2020image,liu2021swin,carion2020end,jiang2024dgpic,zhou2023transvod,he2021end}. Thanks to both self-attention and
cross-attention mechanisms, Transformers have a strong capability in sequence modeling and improving the information interaction between the Transformer encoder and the decoders. Recently, PoinTr~\cite{yu2021pointr} adopted a Transformer encoder-decoder architecture and has been presented to reformulate PCC as a set-to-set translation problem, showing excellent performance within specific domains. %
Unfortunately, no relevant works have investigated the transferability of point Transformer toward various target domains.  To the best of our knowledge, this is the first work that studies the transferability of point Transformer in UDA PCC.

\section{Methodology}
In this section, we concentrate on unsupervised domain adaptation (UDA) for point cloud completion (PCC), where the training dataset comprises paired partial-complete point clouds from the labeled source domain and partial input from the unlabeled target domain. The primary objective is to narrow the domain shift between the source and the target domain. To this end, we present Domain Adaptive Point Transformer (DAPoinTr) framework for UDA point cloud completion that aligns the sequence-wise point features with Transformer encoder-decoder architecture to enhance the adaptability toward the real target domains. As shown in Figure~\ref{fig2},
DAPoinTr consists of three key components: Domain Query-based Feature Alignment (DQFA), Point Token-wise Feature alignment (PTFA), and Voted Prediction Consistency (VPC). Specifically, DQFA is presented to narrow the global domain gaps from the sequence via the presented domain proxy and domain query at the Transformer encoder and decoder, respectively. Secondly, PTFA is proposed to close the local domain shifts by aligning the tokens, \emph{i.e.,} point proxy and dynamic query, at the Transformer encoder and decoder, respectively. Finally, VPC is designed to consider different Transformer decoders as multiple of experts (MoE) to vote for the ensembled prediction and pseudo-label generation.

\subsection{Domain Query-based Feature Alignment}
To ensure domain invariance in the sequence features of point cloud, we propose Dynamic Query-based Feature Alignment (DQFA) to align source and target features from a global perspective. Specifically, on the Transformer encoder side, a \textit{domain proxy} embedding denoted as $q^{enc}_d$ is defined to encode the global context of the sequence, aggregates useful domain-specific features of the entire token sequences, and narrows the global domain gaps in layout and inter-sketch relationships of the point cloud. During the encoding phase, \textit{domain proxy} $q^{enc}_d$ is integrated with point proxy sequences and then fed forwarded to the Transformer encoder for adversarial alignment. Specifically, the initial input $\mathcal{X_E}$ is constructed by concatenating $q^{enc}_d$ with the sequence of point proxy, $f^i$ and then adding the position encoding $Pos^{enc}_d$ generated by the DGCNN. This can be formulated as follows:
\begin{equation}
    \mathcal{X_E}\!=\![q^{enc}_{d};\! f^1\!,\!f^2\!,\!...\!,\!f^N]\! +\! [Pos^{enc}_{d}\!;\! pos^1\!,\!pos^2\!,\!...\!,\!pos^N] 
\end{equation}
To align domain distributions, we feed the domain proxy into the domain-query discriminator and expect that the learned features of the domain proxy cannot be distinguished by the discriminator. $\lambda$ denotes the domain label (0 for the source domain and 1 for the target domain):
\begin{equation}
    \mathcal L_{enc_q}\! =\! \lambda{logD_{enc_q}\!(\mathcal{X_E})}\!+\!(1\!-\!\lambda)log(\!1\!-\!D_{enc_q}\!(\mathcal{X_E}))
\end{equation}
Similarly, a \textit{domain query} $q^{dec}_d$ is presented to encode some sketch relationships for adaptation and fuse the context features from each dynamic query generated by the Query Generator in the sequence. $q^{dec}_d$ is concatenated with dynamic queries to construct $X_D$ before being fed into the Transformer decoder, which can be formulated as follows:
\begin{equation}
    \mathcal {X_D}\!=\![q^{dec}_{d}\!;\! f^{'1}\!,\!f^{'2}\!,\!...\!,\!f^{'N}]\!+\![pos^{'dec}_{d}\!;\! pos^{'1}\!,\!pos{'^2}\!,\!...\!,\!pos^{'N}]
\end{equation} where $f'$ means the global feature and  $pos'$  represents the position embedding generated from Transformer encoder. 
During the decoding phase, we input the domain query to the aforementioned discriminator to align distributions:
\begin{equation}
    \mathcal L_{dec_q}\! =\!\lambda{logD_{dec_q}\!(\mathcal{X_D})}\!+\!(1\!-\!\lambda)log(1-D_{dec_q}\!(\mathcal {X_D}))
\end{equation}

\subsection{Point Token-wise Feature Alignment}
Although the Domain Query-based Feature Alignment (DQFA) can effectively bridge domain gaps from a global perspective, it is less adept at tackling the domain shifts associated with local shape. To address this limitation, we introduce a Point Token-wise Feature Alignment (PTFA) to close the local domain shifts by aligning the tokens at the Transformer encoder and decoder, respectively.

Concretely, Point Proxy Feature Alignment (PPFA) is presented to align each token in the Transformer encoder sequence, \emph{i.e.,} \textit{point proxy}, to alleviate domain gaps caused by local shape, appearance, \emph{etc}. 
Specifically, each point proxy in the Transformer encoder sequence is 
fed into the Point Token-wise Discriminator for adversarial feature alignment:
\begin{equation}
    \mathcal L_{enc_k}\! =\! -\frac{1}{N}\!{\sum_{i=1}^{N}}\!(\lambda{logD_{enc_k}\!(X_p)}\!+\!(1\!-\!\lambda)log(1-D_{enc_k}\!(X_p)))
\end{equation}
 where $X_p$ indicates the point proxies and $\lambda$ indicates domain label (0 for the source domain and 1 for the target domain).

In addition, Dynamic Query Feature Alignment (DQFA) is proposed to align the \textit{dynamic query} at the object levels, thus narrowing the domain gaps on the Transformer decoder side via the Point Token-wise Discriminator $D_{dec_k}$:
\begin{equation}
    \mathcal L_{dec_k}\! =\! -\frac{1}{N}\!{\sum_{i=1}^{N}}\!(\lambda{logD_{dec_k}\!(X_q)}\!+\!(1\!-\!\lambda)log(1-D_{dec_q}\!(X_q)))
\end{equation}
where $X_p$ represents the dynamic query and $\lambda$ indicates domain label (0 for source domain and 1 for  target domain). 

\noindent \textbf{Remarks.} Although the point token-wise sequence feature alignment is adopted on both the encoder and decoder side of point Transformers, it has different implications. Concretely, each token in the Transformer encoder sequence, \emph{i.e.,} point proxy, encodes the local features, and thus Point Proxy Feature Alignment (PPFA) focuses on narrowing the domain gaps caused by that local shape, appearance \emph{etc}. In contrast, each token in the Transformer decoder sequence, \emph{i.e.,}  dynamic queries, captures the local structural information at the object level, and therefore, Dynamic Query Feature Alignment (DQFA) concentrates more on alleviating the domain shifts caused by sketch information.

\subsection{Voted Prediction Consistency}
Due to the various performances of the point cloud predictions at different layers within the Transformer decoder, we introduce Voted Prediction Consistency (VPC) that considers different Transformer decoders as multiple of experts (MoE) to vote for the ensembled prediction, alleviating the domain shifts between the source and target domain.

To be specific, the coordinates $Pred_l$ of the point clouds predicted by each decoder layer are meticulously preserved, and the mean of these coordinates $M_{mean}$ is calculated within the 3D coordinate system. This process enforces a consistency constraint across the outputs of each decoder layer to optimize the quality of the recovered shape. 
\begin{equation}
    \mathcal L_{cons} = CD(M_{mean}, Pred_l)
\end{equation}
In addition, we use the results from the Voted Prediction Consistency as a threshold criterion for selecting high-quality predictions to generate pseudo-labels to further improve the robustness and transferability.

This process serves to further regularize the sequence features, enhancing the discriminability and generalization capabilities of our proposed approach. By systematically refining the quality of pseudo-labels incorporated into the training process, we ensure that only the most representative and accurate predictions are selected to alleviate domain gaps and improve the performance of target samples. Therefore, the total loss function can be formulated as follows:
\begin{equation}
    \mathcal L_{total} = \alpha(\mathcal L_{enc_q}+\mathcal L_{dec_q})+\beta(\mathcal L_{enc_k}+\mathcal L_{dec_k})+\gamma{\mathcal L_{cons}}
\end{equation} 
where $\alpha$, $\beta$ and $\gamma$ are the weights balancing these losses.

\section{Experiments}
% In this section, we first describe the experimental setting in Section~\ref{section4.1}, including the datasets, metrics, and implementation details. Then, in Section~\ref{section4.2}, we demonstrate the effectiveness of our DAPoinTr framework compared to the state-of-the-art approaches on several benchmark datasets. Next, in Section~\ref{section4.3}, we conduct ablation studies to investigate the role of each component in the method. Finally, we provide detailed visualization comparisons in Section~\ref{section4.4}. 

\subsection{Experimental Setting}
\label{section4.1}
\noindent\textbf{Datasets.} Following protocols of previous UDA PCC methods \cite{chen2019unpaired,gong2022optimization}, we use 3D data from CRN~\cite{wang2020cascaded} as the source domain, and the datasets, including Real-World Scans, 3D-FUTURE~\cite{fu20213d} and ModelNet~\cite{wu20153d}, as target domains. As for \textit{CRN} dataset,
we take 26,863 samples of CRN from shared categories between CRN and other datasets for DA. For the benchmark of \textit{Real-World Scans}, we evaluate the performance on ScanNet~\cite{dai2017scannet}, MatterPort3D~\cite{chang2017matterport3d}, and KITTI~\cite{geiger2012we}, where the tables and chairs in ScanNet and MatterPort3D, and cars in KITTI are used for evaluation. We re-sample the input scans to 2, 048 points for unpaired training
and inference to match the virtual dataset. As for \textit{3D-FUTURE}, it only contains indoor furniture models that appear like real objects. We obtain partial shapes and complete ones with 2,048 points from 5 different viewpoints as the target domain. As for \textit{ModelNet}, a subset extracted from the original ModelNet40 dataset~\cite{wu20153d}, 2,048 points are taken for both partial and complete shapes to match CRN, and 6 shared categories between ModelNet40 and CRN are used for evaluation.  

\subsubsection{Implementation Details.} All experiments were conducted on an RTX 4090 with 64GB RAM. We use the refinement module of PoinTr~\cite{yu2021pointr} as the backbone of our PCC network. For the training, we employ the initial learning rate of $2 \times 10^{-4}$ and a weight decay of $5 \times 10^{-5}$. The batch size is set to 2. To balance losses, weights of $\alpha$, $\beta$, and $\gamma$ are set as 0.025, 0.25, and 0.01 respectively. Following previous works, we adopt Unidirectional Chamfer Distance (UCD), Unidirectional Hausdorff Distance (UHD), and Chamfer Distance (CD) as the metrics for evaluation. 

\begin{table}[t]
\centering
\footnotesize
\setlength{\tabcolsep}{1mm}
\begin{tabular}{l|l|cccccc}
\toprule
Methods&Avg&Cabinet&Chair&Lamp&Sofa&Table \\
\midrule
Pcl2Pcl&92.83&57.23&43.91&157.86&63.23&141.92\\
ShapeInv.&53.21 &38.54&26.30&48.57&44.02&108.60\\
Cycle4Co.&45.39 &32.62&34.08&77.19&43.05&40.00\\
ACL-SPC&35.97&70.12&23.87&31.75&28.74&25.38\\
OptDE&28.99&28.37&21.87&29.92&37.98&26.81 \\

DAPoinTr (Ours)&\textbf{22.35} &\textbf{18.46} &\textbf{17.60} &\textbf{27.91} &\textbf{23.08} &\textbf{24.71}\\
\bottomrule
\end{tabular}
\caption{Cross-domain completion results on 3D-FUTURE. We take Chamfer Distance (CD)$\downarrow$ as the metric to evaluate the performance, and the scale factor is $10^4$. Lower is better.}
\label{table1}
\end{table}

\subsection{Comparison Results}
\label{section4.2}
In this section, we conduct extensive experimental comparisons on the widely-used real-world and synthetic benchmarks, including KITTI, ScanNet, and MatterPort for real-world scans, and 3D-FUTURE and ModelNet for synthetic data, to evaluate the performance of our method.

\noindent \textbf{Comparison Methods.} %To evaluate the performance of our approach, 
We compare our method with several state-of-the-art UDA point cloud completion techniques, including Pcl2Pcl~\cite{chen2019unpaired}, ShapeInversion~\cite{zhang2021unsupervised}, Cycle4Completion~\cite{wen2021cycle4completion}, Optde~\cite{gong2022optimization} and ACL-SPC~\cite{hong2023acl}.

\noindent \textbf{Results on 3D FUTURE dataset.}
As illustrated in Table \ref{table1}, our DAPoinTr shows substantial superiority to state-of-the-art UDA PCC approaches on the 3D-FUTURE dataset, achieving significantly better performance across all categories. Notably, we achieved an average improvement of 6.64 in Chamfer Distance (CD) across all categories compared to the 2nd best method OptDE. Additionally, unlike previous UDA PCC methods, which often struggle to bridge domain gaps, especially in cabinet and sofa categories, our results indicate substantial enhancements in these categories, as evidenced by significant margins of improvement (see the 3nd and 6th columns in Table \ref{table1}). This is because our approach can better tackle domain discrepancies by aligning feature distribution between source and target domain, as well as leveraging the capacity of Transformer to capture both global and local features. 

\begin{table}[t]
\centering
\footnotesize
\small
\setlength{\tabcolsep}{0.5mm}
\begin{tabular}{l|l|cccccc}
\toprule
Methods&Avg&Plane&Car&Chair&Lamp&Sofa&Table \\
\midrule
Pcl2Pcl&68.14&18.53&17.54&43.58&126.80&38.78&163.62\\
ShapeInv.& 41.61 &3.78& 15.66& 22.25& 60.42&22.25 & 125.31\\
Cycle4Co.& 28.65 &5.77& 11.85& 26.67& 83.34& 22.82& 21.47 \\
ACL-SPC&34.89&5.75&11.73&43.08&106.29&25.62&16.89\\
OptDE& 15.94 &\textbf{2.18}& 9.80& 14.71& 39.74& 19.43& \textbf{9.75} \\
DAPoinTr (Ours)&\textbf{13.79} &2.38 &\textbf{8.04} &\textbf{13.83} &\textbf{33.26} &\textbf{12.72} &12.51\\
\bottomrule
\end{tabular}
%}
\caption{Cross-domain completion results on ModelNet. We take Chamfer Distance (CD)$\downarrow$ as the metric to evaluate the performance, and the scale factor is $10^4$. Lower is better.}
\label{table2}
\end{table}

\begin{table}[t]
\centering
\footnotesize
\setlength{\tabcolsep}{0.5mm}
\begin{tabular}{@{}l|cc|cc|cc@{}}
\toprule
\multirow{2}{*}{Methods}
&\multicolumn{2}{c}{ScanNet}&\multicolumn{2}{c}{MatterPort3D}&\multicolumn{2}{c}{KITTI}\\
\cmidrule(lr){2-7}
&Chair&Table&Chair&Table&Car \\

\midrule
Pcl2Pcl&17.3/10.1&9.1/11.8&15.9/10.5&6.0/11.8&9.2/14.1\\
ShapeInv.&3.2/10.1 &3.3/11.9&3.6/10.0&3.1/11.8&2.9/13.8\\
Cycle4Co.&5.1/6.4 &3.6/5.9&8.0/8.4&4.2/6.8&3.3/5.8\\
ACL-SPC&1.4/4.7& 1.8/5.1&1.8/4.8&2.1/4.9&2.0/4.9&\\
OptDE&2.6/5.5&1.9/4.6&3.0/5.5&1.9/5.3&1.6/3.5 \\
DAPoinTr (Ours)&\textbf{1.1/2.7} &\textbf{0.96/2.7} &\textbf{1.3/2.9} &\textbf{1.2/2.8} &\textbf{0.45/1.8} \\
\bottomrule
\end{tabular}
\caption{Cross-domain completion results on Real-World Scans. We take [UCD$\downarrow$/UHD$\downarrow$] as the metric to evaluate the performance, and the scale factor is $10^4$ and $10^2$ for UCD and UHD, respectively. UCD: Unidirectional Chamfer Distance. UHD: Unidirectional Hausdorff Distance. A lower value of UCD or UHD is better. }
\label{table3}
\end{table}

% ablation result
\begin{table*}[t]
\centering
\footnotesize
\begin{tabular}{l|ccc|ccccc|c}
\toprule
Methods&PTFA&DQFA&VPC&Cabinet&Chair&Lamp&Sofa&Table&Avg \\
\midrule
Baseline* (PoinTr)&&&&32.88&24.21&41.76&33.76&35.27&33.58\\
\midrule
Our DAPoinTr*&\checkmark&\checkmark&\checkmark&31.62&20.01&29.01&32.97&28.72&28.47\\
\midrule
Baseline (PoinTr w SPD)&&&&20.82&21.68&29.43&24.68&32.09&25.74\\
\midrule
\multirow{4}{*}{Our DAPoinTr}

&\checkmark&&&20.65&18.77&32.26&24.39&26.67&24.55 \\
&&\checkmark&&20.37&18.24&29.80&25.02&26.13&23.91 \\
&\checkmark&\checkmark&&20.21&18.11&29.13&23.84&25.01&23.26 \\

&\checkmark&\checkmark&\checkmark&\textbf{18.46} &\textbf{17.60}&\textbf{ 27.91}&\textbf{23.08}&\textbf{24.71}&\textbf{22.35} \\

\bottomrule
\end{tabular}
\caption{Ablation studies on each proposed module of our DAPoinTr on 3D-FUTURE dataset. We take Chamfer Distance (CD)$\downarrow$ as the metric to evaluate the performance, and the scale factor is $10^4$. Lower is better.}
\label{table4}
\end{table*}

% ModelNet Result Visualization
\begin{figure*}[t!]
\centering
\includegraphics[width=0.9\textwidth]{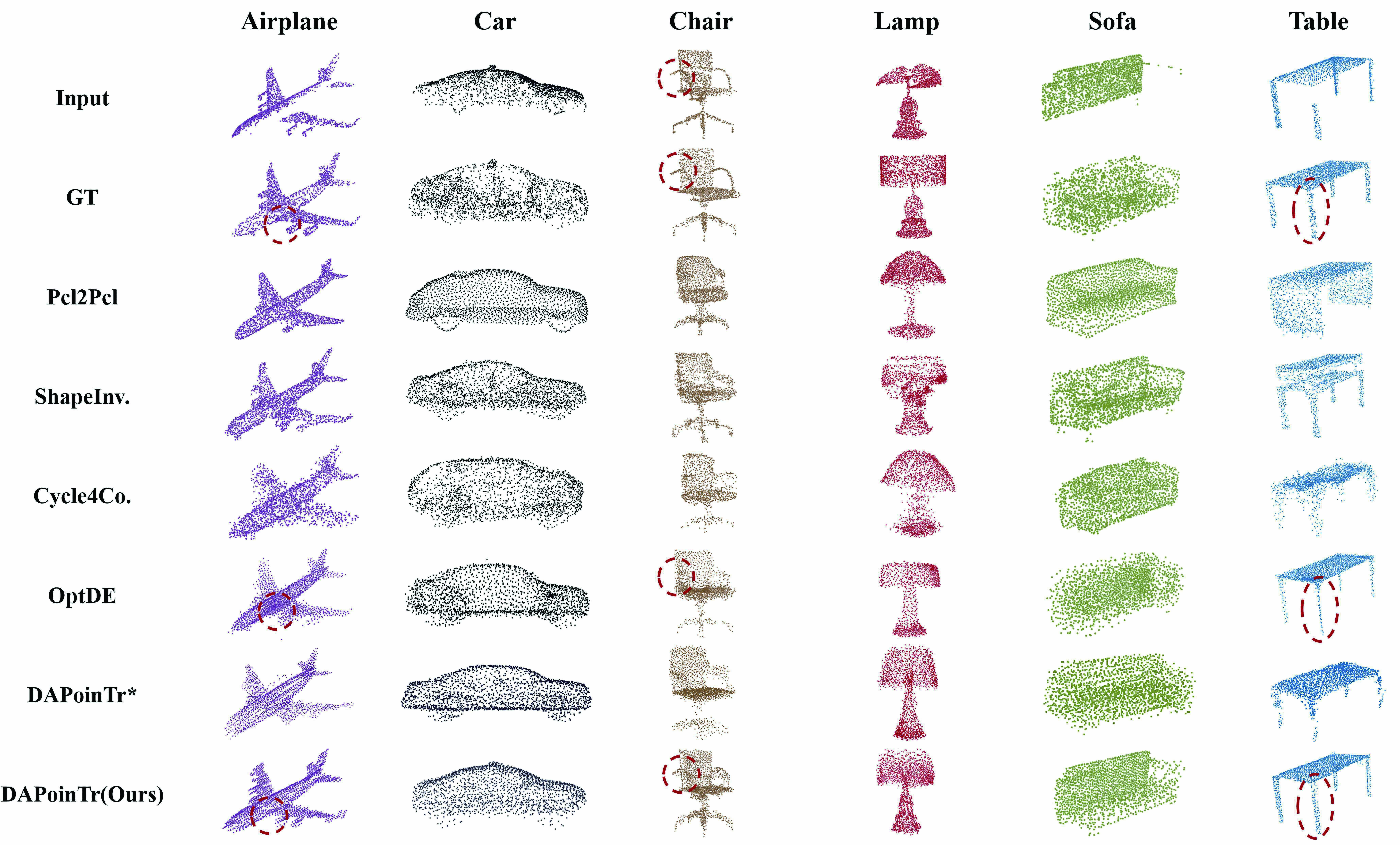} 
\caption{Visualization comparisons with state-of-the-art PCC methods on ModelNet dataset. }
\label{fig3}
\end{figure*}

\noindent \textbf{Results on ModelNet dataset.}
Table \ref{table2} demonstrates that our method outperforms state-of-the-art methods in most categories on ModelNet. Specifically, DAPoinTr achieves a remarkable improvement in the category of Lamp and Sofa in terms of the mean Chamfer Distance. This superiority indicates that the innovative designs of our model can effectively alleviate domain shifts and close domain gaps at the object level. This remarkable improvement underscores the robustness and adaptability of the innovative design elements incorporated within our model.

\noindent \textbf{Results on Real-World Scans.}
 In this benchmark, we utilize the UCD and UHD as evaluation metrics. As shown in Table \ref{table3}, it is obvious that our approach outperforms previous methods across all real-world scan datasets in terms of UCD and UHD metrics by a large margin, which indicates our model can better maintain the overall geometry of partial inputs and effectively align the feature distribution and bridge the domain gaps in output predictions. 

\subsection{Ablation Studies}
\label{section4.3}
In this section, we conduct ablation studies to verify the contribution of each proposed module, as shown in Table \ref{table4}. In all ablation experiments, we use the CRN dataset as the source domain and 3D-FUTURE as the target domain.

\begin{figure*}[ht!]
\centering
\includegraphics[width=0.9\textwidth]{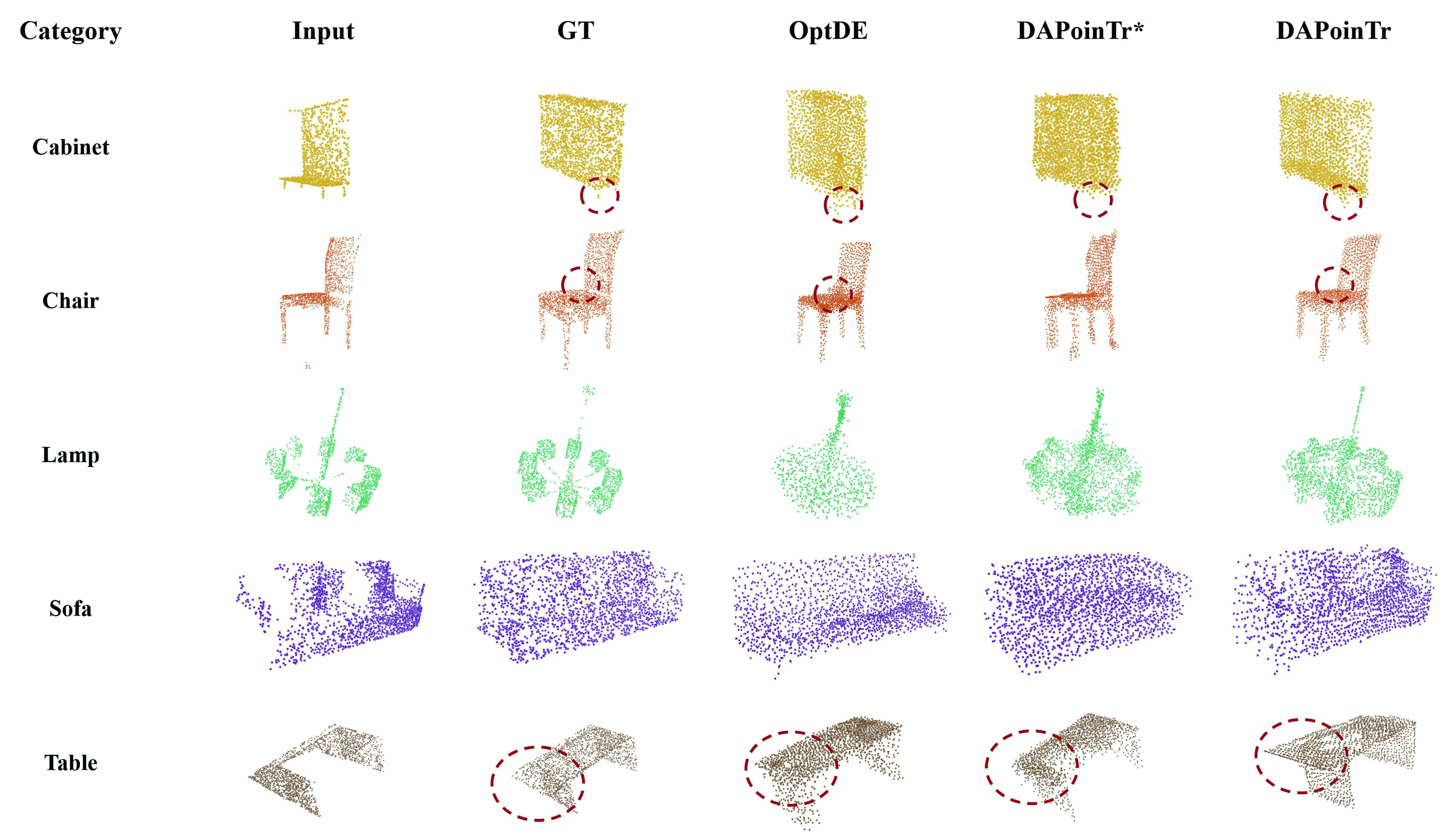} % Reduce the figure size so that it is slightly narrower than the column.
\caption{Visualization comparisons with state-of-the-art PCC methods on 3D-FUTURE dataset. }
\label{fig4}
\end{figure*}

Initially, we employed the original PoinTr~\cite{yu2021pointr} (denoted as Baseline*) to train on the source domain and directly tested it on the target domain. However, we observe that the original PoinTr produces undesirable performance (CD: 33.58) on the target domain due to the lack of learning domain-invariant sequence features.
In contrast, by adding our proposed components into Baseline*, our DAPoinTr*  significantly improves transferability and manages to achieve a better completion performance (CD: 28.47). 

Considering that PoinTr employs FoldingNet~\cite{yang2018foldingnet} to decode the recovered complete point cloud, we observed that FoldingNet struggles with effective recovery when handling complex shapes. Inspired by SnowFlakeNet~\cite{xiang2021snowflakenet}, we replace the FoldingNet in PoinTr with the SPD decoder from SnowFlakeNet  (denoted as Baseline) to reconstruct the complete point cloud. The results in the 3rd row of Table~\ref{table4} confirm that the SPD decoder demonstrates better performance, making it a competent solution for complex geometry. 

Based on this superior baseline, we conduct detailed ablation studies to reveal the impact of each proposed component.
By adding PTFA into the baseline, we effectively decrease the average performance from 25.74 to 24.55, demonstrating the effectiveness of narrowing local domain shifts. By further adding DQFA, we achieve 23.91 of the average CD performance, demonstrating the effectiveness in narrowing global domain shifts. Besides, when using both PTFA and DQFA, the results are much better than using merely one. In challenging categories such as sofas and chairs, as shown in Table~\ref{table4}, the integration of PTFA and DQFA effectively diminishes the impact of domain gaps and promotes the learning of domain-invariant sequence features.  This is because each module is indispensable in enhancing the model's proficiency in capturing both local details and global structures, and in boosting the model's adaptability in the realm of UDA PCC. 
Finally, when further adding VPC, the average CD result is further decreased to 22.35. The improvement is particularly notable in the categories of cabinets and tables, which show significant enhancements over previous UDA PCC methods. These results confirm that these individual components are complementary and together they significantly promote the performance. 

\subsection{Visualization and Analysis}
\label{section4.4}

\noindent \textbf{Visualization of Completed Point Clouds on ModelNet}.
As depicted in Figure \ref{fig3}, we visualize the qualitative results on ModelNet. When comparing the visualization results of our DAPoinTr with the previous state-of-the-art method, OptDE, our approach consistently generates more accurate completions with superior fine-grained local details. This improvement is particularly evident in the cases of chairs and tables, as highlighted in the second and final rows of the figure. These results demonstrate the enhanced capability of our method in rendering more detailed and realistic structures in complex object categories. 

\noindent \textbf{Visualization of Completed Point Clouds on 3D-FUTURE}.
Figure \ref{fig4} visualizes completed point clouds on the 3D-FUTURE dataset.  Our approach excels in capturing both local details and global structures. Our DAPoinTr consistently reconstructs more accurate complete shapes, maintaining superior structural integrity and a more even distribution of points compared to other state-of-the-art methods. This demonstrates our robustness to delivering better reconstructions across various complex scenarios.

\section{Conclusion}
In this paper, we introduce Domain Adaptive Point Transformer (DAPoinTr), a novel and pioneering framework aimed at enhancing the transferability of point Transformers (PoinTr) for point cloud completion across varying domains. Through the integration of several new modules including Domain Query-based Feature Alignment (DQFA), Point Token-wise Feature Alignment (PTFA), and Voted Prediction Consistency (VPC), our DAPoinTr effectively addresses both global and local domain discrepancies, resulting in superior structural integrity and point distribution. 
Extensive experiments on the synthetic and real datasets demonstrate remarkable improvements over existing methods, particularly in complex object categories.  

%\appendix
\bibliography{aaai25}

% \title{Supplementary Material of DAPoinTr: Domain Adaptive Point \\Transformer for Point Cloud Completion}

\twocolumn[
\begin{@twocolumnfalse}
	\section*{\centering{DAPoinTr: Domain Adaptive Point Transformer for Point Cloud Completion
	\\Supplementary Material\\[90pt]}}
\end{@twocolumnfalse}
]

% \section{Appendix.}

The supplementary material presents further details, which are structured as follows:

\begin{description}
\small
    %\item[Section 1]: Implementation Details
    \item[Section 6]: Visualization and Analysis
    \begin{description}
        \item[Section 6.1]: TSNE Visualization of PoinTr and DAPoinTr
        \item[Section 6.2]: Visualization Results of Completed Point Clouds
    \end{description}
    \item[Section 7]: Ablation Study of DQFA and PTFA
    \item[Section 8]: Limitations and Future Work
    %\item[Section 4]: Compuation Efficiency
\end{description}
\section{Visualization and Analysis}

\subsection{TSNE Visualization of PoinTr and DAPoinTr}
The  TSNE visualization (Figure \ref{fig1}) compares the feature alignment between the source and target domains for the two different models: PoinTr \cite{yu2021pointr} and our method DAPoinTr. The TSNE result illustrates the feature alignment results from the Transformer encoder of both models. 
As we can see from Figure \ref{fig1}, the TSNE plots for PoinTr exhibit poor alignment between the source and target domain. It is clearly noticed from the separation and lack of overlap between the red (source domain) and blue (target domain) points. The distinct clustering indicates that the features learned by PoinTr do not generalize well across domains, resulting in ineffective feature alignment. In contrast, our DAPoinTr model, which involves the proposed modules, including PTFA and DQFA, shows significantly better alignment between the source and target domains. The TSNE plots reveal much more overlap between the red and blue points, indicating that the feature distributions from both domains are more closely aligned. This demonstrates our DAPoinTr is more effective at bridging the domain gap, leading to a more robust and generalizable model. 
\begin{figure}[t]
    \centering
    \includegraphics[width=1.0\columnwidth]{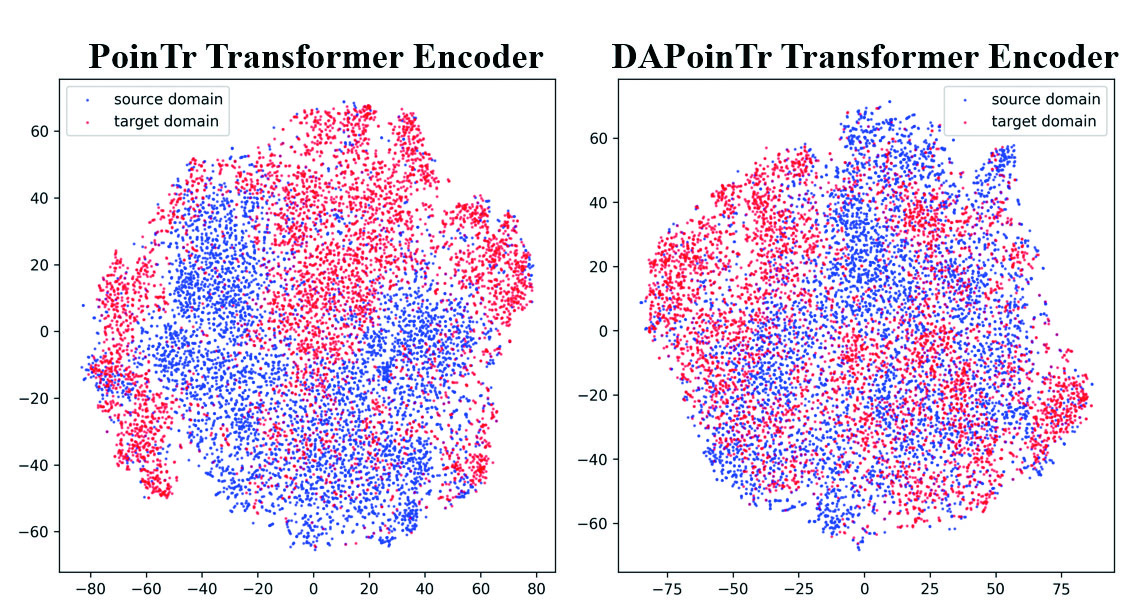} % Reduce the figure size so that it is slightly narrower than the column.
    \caption{TSNE Visualization of feature distribution for PoinTr and DAPoinTr. }
    \label{fig1}
\end{figure}

% ablation result
\begin{table*}[t]
\centering
\footnotesize
\small
\setlength{\tabcolsep}{0.5mm}
\begin{tabular}{l|c|c|c|c|c|c}
\toprule
Methods&Domain Proxy Alignment&Point Proxy Alignment&Domain Query Alignment&Dynamic Query Alignment&VPC&Chair\\
&(DQFA)&(PTFA)&(DQFA)&(PTFA)&&\\
\midrule
Baseline&&&&&&21.68\\  
\midrule
\multirow{6}{*}{DAPoinTr}
&&\checkmark&\checkmark&\checkmark&&18.16 \\
&\checkmark&&\checkmark&\checkmark&&18.32 \\
%&&\checkmark&\checkmark&&20.37&18.11&29.13&24.39&25.80&23.56 \\
&\checkmark&\checkmark&&\checkmark&&18.28 \\

&\checkmark&\checkmark&\checkmark&&&18.39\\
&\checkmark&\checkmark&\checkmark&\checkmark&&18.11\\
&\checkmark&\checkmark&\checkmark&\checkmark&\checkmark&\textbf{17.60}\\
\bottomrule
\end{tabular}
%}
%\captionsetup{width=\textwidth}
\caption{Ablation studies on each module of our DAPoinTr on the category of Chair of 3D-FUTURE dataset. We take Chamfer Distance (CD)$\downarrow$ as the metric to evaluate the performance, and the scale factor is $10^4$. Lower is better.}
\label{table1}
\end{table*}

\subsection{Visualization Results of Completed Point Clouds}
\noindent \textbf{Visualization of Completed Point Clouds on MatterPort3D.} 
In here, we compared the visualization results of the MatterPort3D \cite{chang2017matterport3d} dataset with previous state-of-the-art methods, including Pcl2Pcl~\cite{chen2019unpaired}, ShapeInversion~\cite{zhang2021unsupervised}, Cycle4Completion~\cite{wen2021cycle4completion}, and OptDE~\cite{gong2022optimization}. As depicted in Figure \ref{fig2}, it can be easily observed that our method can capture local details and generate more precise shapes. This superior performance is attributed to the innovative point token-wise alignment and domain query-based alignment,  ensuring that our model better recovers local geometric structures and global shapes.

\noindent \textbf{Visualization of Completed Point Clouds on KITTI.}
Figure \ref{fig3} illustrates the visualization of completed point cloud results on the real-world dataset KITTI \cite{geiger2012we}. This visualization clearly shows the superiority of our model compared with other methods in terms of domain adaptation and the preservation of structural integrity in point cloud completion. 

\begin{figure*}[htbp]
\centering
\includegraphics[width=1.0\textwidth]{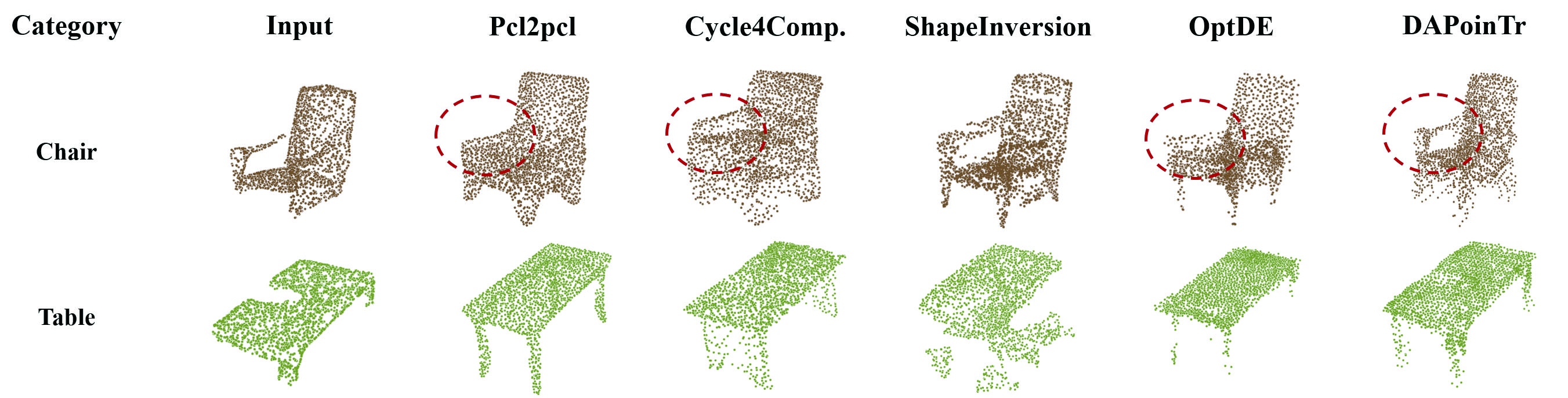} % Reduce the figure size so that it is slightly narrower than the column.
\caption{Visualization of completed point cloud on the real-world dataset MatterPort3D.}
\label{fig2}
\end{figure*}

\begin{figure*}[htbp!]
\centering
\includegraphics[width=1.0\textwidth]{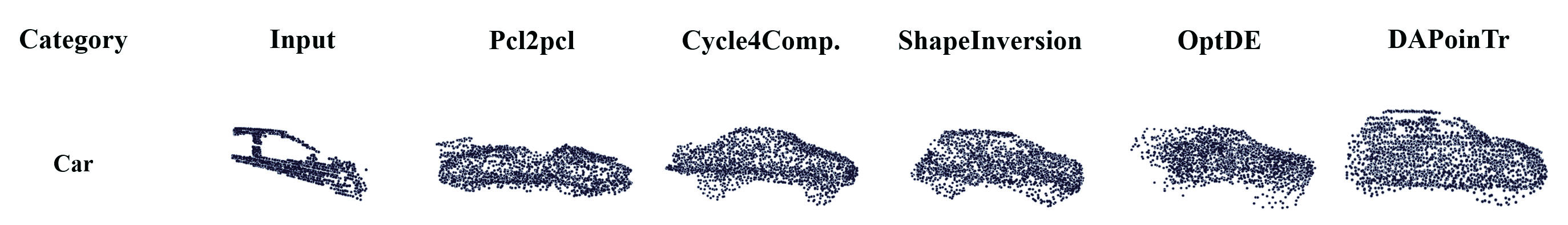} % Reduce the figure size so that it is slightly narrower than the column.
\caption{Visualization of completed point cloud on the real-world dataset KITTI.
}
\label{fig3}
\end{figure*}

\begin{figure}[htbp!]
    \centering
    \includegraphics[width=1.0\columnwidth]{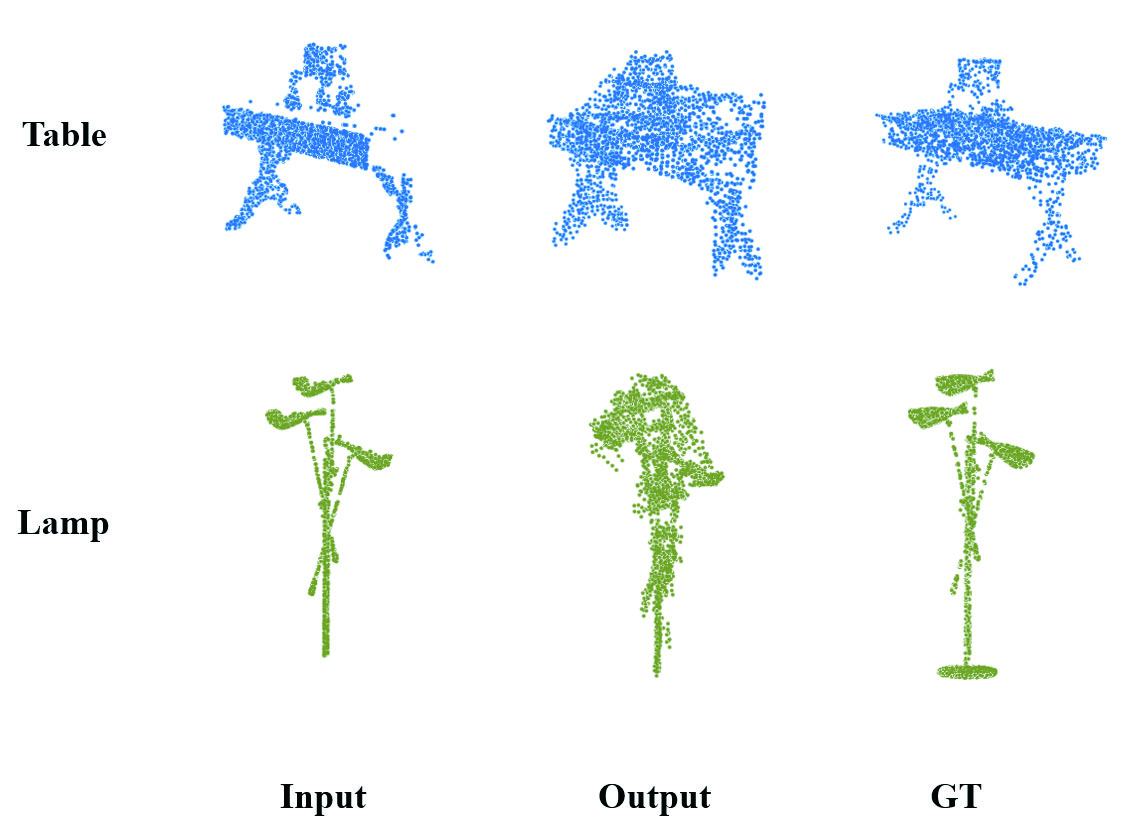}
    \caption{Failure cases.}
    \label{fig4}
\end{figure}

\section{Ablation Study of DQFA and PTFA}
%In our ablation studies for our designed approach, DAPoinTr, 
We utilize the CRN dataset as the source and the 3D-FUTURE dataset as the target domain to assess the impact of each proposed component on the performance in PCC across domains. We selected the chair class to demonstrate the influence of each module. As depicted in Table \ref{table1}, the baseline model, without any adaptive enhancements, recorded a Chamfer Distance (CD) of 21.68. 
Initially, we conducted the experiment without individual components, including Domain Proxy, Point Proxy, Domain Query, and Dynamic Query, and the evaluation results are 18.16, 18.32, 18.28, and 18.39, respectively. Experimental results proved the important roles of the proposed components for feature alignment and alleviating domain gaps compared with the baseline. Integrated with all designed components of PTFA (Point Proxy and Dynamic Query) and DQFA (Domain Proxy and Domain Query) further improved the performance (CD decreased to 18.11), indicating the effectiveness in capturing domain-invariant features for better completion. 
The full integration of all components, including VPC, enabled the lowest CD of 17.60, underscoring the effectiveness of ensemble predictions and consistency checks in refining the final model outputs. Our approach not only minimized domain discrepancies but also remarkably enhanced the precision and adaptability. The systematic inclusion of each module illustrated that local and global feature alignments are essential for achieving optimal performance in point cloud completion across varied domains.

\section{Limitations and Future Work}
Our DAPoinTr demonstrates elegant and successful integration of Transformer architecture into unsupervised point cloud completion, achieving state-of-the-art performance. 
While our approach significantly improves the recovery of local and global details from partial inputs, it still faces considerable domain gaps when handling specific objects that have complex structures. As illustrated in Figure \ref{fig4}, there are failure cases when dealing with more complex object structures. In future work, we would like to explore this.

\label{sec:reference_examples}

\end{document}